%% file: fractalnet.tex
\documentclass{article}

\usepackage{iclr2017_conference,times}
\definecolor{darkblue}{rgb}{0.0,0.0,0.55}
\usepackage[
   pagebackref=false,
   breaklinks=true,
   letterpaper=true,
   colorlinks,
   citecolor=darkblue,
   bookmarks=false]{hyperref}
\usepackage{url}

\usepackage[utf8]{inputenc} 
\usepackage[T1]{fontenc}    
\usepackage{url}            
\usepackage{booktabs}       
\usepackage{amsfonts}       
\usepackage{nicefrac}       
\usepackage{microtype}      
\usepackage{capt-of}
\usepackage{cleveref}
\usepackage{comment}

\usepackage{graphicx}
\usepackage{arydshln}
\usepackage{mathabx}
\usepackage{tikz}
\usepackage{standalone}
\usepackage{ifthen}
\usetikzlibrary{arrows}
\usetikzlibrary{calc}
\usetikzlibrary{backgrounds}
\usetikzlibrary{shapes.arrows}

\input{include/macros}

\title{%
   FractalNet:\\%
   \resizebox{\textwidth}{!}{Ultra-Deep Neural Networks without Residuals}%
}

\author{%
   Gustav Larsson\\
   University of Chicago\\
   \texttt{larsson@cs.uchicago.edu}\\
   \And
   Michael Maire\\
   TTI Chicago\\
   \texttt{mmaire@ttic.edu}\\
   \And
   Gregory Shakhnarovich\\
   TTI Chicago\\
   \texttt{greg@ttic.edu}%
}

\iclrfinalcopy

\begin{document}

\maketitle

\begin{abstract}
\input{include/sec0-abstract}
\end{abstract}

\section{Introduction}
\label{sec:intro}
\input{include/sec1-introduction}

\section{Related Work}
\label{sec:related}
\input{include/sec2-related}

\section{Fractal Networks}
\label{sec:method}
\input{include/sec3-method}

\section{Experiments}
\label{sec:results}
\input{include/sec4-results}

\section{Conclusion}
\label{sec:discussion}
\input{include/sec5-final}

\subsection*{Acknowledgments}
We gratefully acknowledge the support of NVIDIA Corporation with the donation
of GPUs used for this research.  This work was partially supported by the NSF
award RI:1409837.

{\small
\bibliographystyle{iclr2017_conference}
\bibliography{fractalnet}
}

\end{document}

%% file: include/macros.tex
\def\eg{\emph{e.g.}}

\newcommand{\fracnet}{FractalNet}

\newcommand{\resnet}{ResNet}
\newcommand{\resnets}{ResNets}
\newcommand{\dropout}{dropout}
\newcommand{\dropconn}{drop-connect}
\newcommand{\droppath}{drop-path}

\newcommand{\Fracnet}{FractalNet}

\newcommand{\Resnet}{ResNet}

\newcommand{\Dropout}{Dropout}

\newcommand{\Droppath}{Drop-path}

%% file: include/sec0-abstract.tex
We introduce a design strategy for neural network macro-architecture based on
self-similarity.  Repeated application of a simple expansion rule generates
deep networks whose structural layouts are precisely truncated fractals.  These
networks contain interacting subpaths of different lengths, but do not include
any pass-through or residual connections; every internal signal is transformed
by a filter and nonlinearity before being seen by subsequent layers.  In
experiments, fractal networks match the excellent performance of standard
residual networks on both CIFAR and ImageNet classification tasks, thereby
demonstrating that residual representations may not be fundamental to the
success of extremely deep convolutional neural networks.  Rather, the key may
be the ability to transition, during training, from effectively shallow to
deep.  We note similarities with student-teacher behavior and develop
drop-path, a natural extension of dropout, to regularize co-adaptation of
subpaths in fractal architectures.  Such regularization allows extraction of
high-performance fixed-depth subnetworks.  Additionally, fractal networks
exhibit an anytime property: shallow subnetworks provide a quick answer, while
deeper subnetworks, with higher latency, provide a more accurate answer.

%% file: include/sec1-introduction.tex
Residual networks~\citep{he2015deep}, or {\resnets}, lead a recent and dramatic
increase in both depth and accuracy of convolutional neural networks,
facilitated by constraining the network to learn residuals.  {\Resnet}
variants~\citep{he2015deep,he2016identity,huang2016stochasticdepth} and related
architectures~\citep{srivastava2015highway} employ the common technique of
initializing and anchoring, via a pass-through channel, a network to the
identity function.  Training now differs in two respects.  First, the objective
changes to learning residual outputs, rather than unreferenced absolute
mappings.  Second, these networks exhibit a type of deep supervision~\citep{
lee2014deeply}, as near-identity layers effectively reduce distance to the
loss.  \cite{he2015deep} speculate that the former, the residual formulation
itself, is crucial.

\begin{figure}[ht]
   \begin{center}
        \begingroup
        \input{#include/fig-overview}%
        \endgroup
    
   \end{center}
   \vspace{-0.02\linewidth}
   \caption{
      \textbf{Fractal architecture.}
      \emph{Left:}
         A simple expansion rule generates a fractal architecture with $C$
         intertwined columns.  The base case, $f_1(z)$, has a single layer
         of the chosen type (\eg~convolutional) between input and output.
         Join layers compute element-wise mean.
      \emph{Right:}
         Deep convolutional networks periodically reduce spatial resolution
         via pooling.  A fractal version uses $f_C$ as a building block
         between pooling layers.  Stacking $B$ such blocks yields a network
         whose total depth, measured in terms of convolution layers, is
         $B \cdot 2^{C-1}$.  This example has depth $40$ ($B=5$, $C=4$).
   }
   \label{fig:fractalnet}
\end{figure}
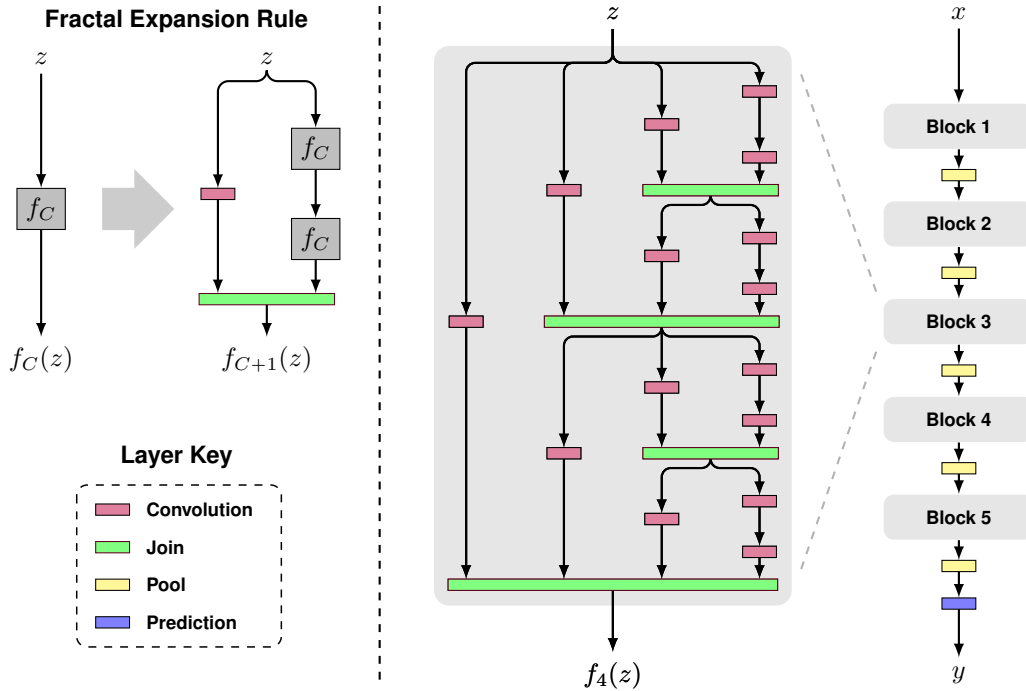

We show otherwise, by constructing a competitive extremely deep architecture
that does not rely on residuals.  Our design principle is pure enough to
communicate in a single word, fractal, and a simple diagram
(Figure~\ref{fig:fractalnet}).  Yet, fractal networks implicitly recapitulate
many properties hard-wired into previous successful architectures.  Deep
supervision not only arises automatically, but also drives a type of
student-teacher learning~\citep{ba2014dodeep,urban2016dodeepsfollowup} internal
to the network.  Modular building blocks of other designs~\citep{
szegedy2015inception,liao2015competitive} resemble special cases of a fractal
network's nested substructure.

For fractal networks, simplicity of training mirrors simplicity of design.
A single loss, attached to the final layer, suffices to drive internal
behavior mimicking deep supervision.  Parameters are randomly initialized.
As they contain subnetworks of many depths, fractal networks are robust to
choice of overall depth; make them deep enough and training will carve out a
useful assembly of subnetworks.

The entirety of emergent behavior resulting from a fractal design may erode
the need for recent engineering tricks intended to achieve similar effects.
These tricks include residual functional forms with identity initialization,
manual deep supervision, hand-crafted architectural modules, and
student-teacher training regimes.  Section~\ref{sec:related} reviews this
large body of related techniques.  Hybrid designs could certainly integrate
any of them with a fractal architecture; we leave open the question of the
degree to which such hybrids are synergistic.

Our main contribution is twofold:
\begin{itemize}
   \item{
      We introduce {\fracnet}, the first simple alternative to {\resnet}.
      {\Fracnet} shows that explicit residual learning is not a requirement
      for building ultra-deep neural networks.
   }
   \item{
      Through analysis and experiments, we elucidate connections between
      {\fracnet} and an array of phenomena engineered into previous deep
      network designs.
   }
\end{itemize}

As an additional contribution, we develop {\droppath}, a novel regularization
protocol for ultra-deep fractal networks.  Without data augmentation, fractal
networks, trained with {\droppath} and {\dropout}~\citep{dropout}, exceed the
performance of residual networks regularized via stochastic depth~\citep{
huang2016stochasticdepth}.  Though, like stochastic depth, it randomly removes
macro-scale components, {\droppath} further exploits our fractal structure
in choosing which components to disable.

{\Droppath} constitutes not only a regularization strategy, but also provides
means of optionally imparting fractal networks with anytime behavior.  A
particular schedule of dropped paths during learning prevents subnetworks of
different depths from co-adapting.  As a consequence, both shallow and deep
subnetworks must individually produce correct output.  Querying a shallow
subnetwork thus yields a quick and moderately accurate result in advance of
completion of the full network.

Section~\ref{sec:method} elaborates the technical details of fractal networks
and {\droppath}.  Section~\ref{sec:results} provides experimental comparisons
to residual networks across the CIFAR-10, CIFAR-100~\citep{CIFAR}, SVHN~\citep{
SVHN}, and ImageNet~\citep{deng2009imagenet} datasets.  We also evaluate
regularization and data augmentation strategies, investigate subnetwork
student-teacher behavior during training, and benchmark anytime networks
obtained using {\droppath}.  Section~\ref{sec:discussion} provides synthesis.
By virtue of encapsulating many known, yet seemingly distinct, design
principles, self-similar structure may materialize as a fundamental component
of neural architectures.

%% file: include/fig-overview.tex
\begin{tikzpicture}[->,>=latex,auto,node distance=3cm,transform shape,
 thick,main node/.style={circle,draw,font=\sffamily\Large\bfseries}]
\def\blockheight{7.0}
\def\layerwidth{0.3}
\def\layerheight{0.04}
\def\columnwidth{1.3}
\def\one{5}

\def\blocks{1}
\def\columns{4}

\def\arrowbuf{0.12}
\tikzstyle{conv}=[rectangle,draw,thick,fill=purple!50,thin,minimum width=1.5cm,minimum height=0.5cm,scale=0.3]
\tikzstyle{pool}=[rectangle,draw,thick,fill=yellow!50,thin,minimum width=1.5cm,minimum height=0.5cm,scale=0.3]
\tikzstyle{dense}=[rectangle,draw,thick,fill=blue!50,thin,minimum width=1.5cm,minimum height=0.5cm,scale=0.3]
\tikzstyle{func}=[rectangle,draw,thick,fill=gray!50,thin,scale=1.0]
\tikzstyle{joiner}=[rectangle,thick,draw,fill=green!50,draw=black!50!purple,thin,minimum width=1.5cm,minimum height=0.5cm,scale=0.3]
\tikzstyle{joiner1}=[rectangle,thick,draw,fill=green!50,draw=black!50!purple,thin,minimum width=14.6cm,minimum height=0.5cm,scale=0.3]
\tikzstyle{joiner2}=[rectangle,thick,draw,fill=green!50,draw=black!50!purple,thin,minimum width=10.4cm,minimum height=0.5cm,scale=0.3]
\tikzstyle{joiner3}=[rectangle,thick,draw,fill=green!50,draw=black!50!purple,thin,minimum width=6.0cm,minimum height=0.5cm,scale=0.3]
\tikzstyle{myround}=[rounded corners=0.12cm]
\tikzstyle{arrow}=[->,thick]

\foreach \b in {1, \blocks} {
    \node[joiner1] (joiner-\b-1-1) at (\one+\columnwidth*\columns/2, -\b*\blockheight) {};
    \foreach \j in {1, ..., 1} {
        \node[joiner2] (joiner-\b-2-\j) at (\one+\columnwidth*2.5, -\b*\blockheight +\blockheight -\blockheight/2 - \j*\blockheight/1 + \blockheight/1) {};
    }
    \foreach \j in {1, ..., 2} {
        \node[joiner3] (joiner-\b-3-\j) at (\one+\columnwidth*3, -\b*\blockheight +\blockheight -\blockheight/4 - \j*\blockheight/2 + \blockheight/2) {};
    }
    \foreach \c/\layers in {1/1, 2/2, 3/4, 4/8} {
        \def\x{\one+\c*\columnwidth - \columnwidth/2}
        \def\y{\blockheight-\b*\blockheight}
        \def\convradius{\blockheight/\layers}
        \foreach \n in {1, ..., \layers} {
            \ifthenelse{\c=4 \AND \n=3}{
                \def\yn{\y - \n*\convradius + 0.5*\convradius - 0.2}
            }{
                \ifthenelse{\c=4 \AND \n=5}{
                    \def\yn{\y - \n*\convradius + 0.5*\convradius - 0.2}
                }{
                    \ifthenelse{\c=4 \AND \n=7}{
                        \def\yn{\y - \n*\convradius + 0.5*\convradius - 0.2}
                    }{
                        \def\yn{\y - \n*\convradius + 0.5*\convradius}
                    }
                }
            }
            \def\xn{\x}
            \node[conv] (conv-\b-\c-\n) at (\xn, \yn) {};
        }

    }

    \node (entry) at (\one+\columnwidth*\columns/2, 0.6) {$z$};
    \node (below-entry) at (\one+\columnwidth*\columns/2, 0.1) {};

    \draw[arrow,myround] (entry.south) to ($(entry.south) + (0, -0.45)$) to ($(entry.south) + (+\columnwidth*1.5, -0.45)$) to  (conv-\b-4-1);
    \draw[arrow,myround] (entry.south) to ($(entry.south) + (0, -0.45)$) to ($(entry.south) + (+\columnwidth*0.5, -0.45)$) to  (conv-\b-3-1);
    \draw[arrow,myround] (entry.south) to ($(entry.south) + (0, -0.45)$) to ($(entry.south) + (-\columnwidth*0.5, -0.45)$) to  (conv-\b-2-1);
    \draw[arrow,myround] (entry.south) to ($(entry.south) + (0, -0.45)$) to ($(entry.south) + (-\columnwidth*1.5, -0.45)$) to  (conv-\b-1-1);

    \draw[arrow] (conv-\b-4-1) to (conv-\b-4-2);
    \draw[arrow] (conv-\b-4-2) to ($(joiner-\b-3-1.north) + (\columnwidth/2, 0)$);
    \draw[arrow,myround] (joiner-\b-3-1.south) to ($(joiner-\b-3-1.south) + (0, -\arrowbuf)$) to ($(joiner-\b-3-1.south) + (\columnwidth/2, -\arrowbuf)$) to (conv-\b-4-3);
    \draw[arrow] (conv-\b-4-3) to (conv-\b-4-4);
    \draw[arrow] (conv-\b-4-4) to ($(joiner-\b-2-1.north) + (\columnwidth, 0)$);
    \draw[arrow,myround] (joiner-\b-2-1.south) to ($(joiner-\b-2-1.south) + (0, -\arrowbuf)$) to ($(joiner-\b-2-1.south) + (\columnwidth, -\arrowbuf)$) to (conv-\b-4-5);
    \draw[arrow] (conv-\b-4-5) to (conv-\b-4-6);
    \draw[arrow] (conv-\b-4-6) to ($(joiner-\b-3-2.north) + (\columnwidth/2, 0)$);
    \draw[arrow,myround] (joiner-\b-3-2.south) to ($(joiner-\b-3-2.south) + (0, -\arrowbuf)$) to ($(joiner-\b-3-2.south) + (\columnwidth/2, -\arrowbuf)$) to (conv-\b-4-7);
    \draw[arrow] (conv-\b-4-7) to (conv-\b-4-8);
    \draw[arrow] (conv-\b-4-8) to ($(joiner-\b-1-1.north) + (\columnwidth*1.5, 0)$);

    \draw[arrow] (conv-\b-3-1) to ($(joiner-\b-3-1.north) - (\columnwidth/2, 0)$);
    \draw[arrow,myround] (joiner-\b-3-1.south) to ($(joiner-\b-3-1.south) + (0, -\arrowbuf)$)  to ($(joiner-\b-3-1.south) + (-\columnwidth/2, -\arrowbuf)$)  to (conv-\b-3-2);
    \draw[arrow] (conv-\b-3-2) to (joiner-\b-2-1);
    \draw[arrow] (joiner-\b-2-1) to (conv-\b-3-3);
    \draw[arrow] (conv-\b-3-3) to ($(joiner-\b-3-2.north) - (\columnwidth/2, 0)$);
    \draw[arrow,myround] (joiner-\b-3-2.south) to ($(joiner-\b-3-2.south) + (0, -\arrowbuf)$) to ($(joiner-\b-3-2.south) + (-\columnwidth/2, -\arrowbuf)$) to (conv-\b-3-4);
    \draw[arrow] (conv-\b-3-4) to ($(joiner-\b-1-1.north) + (\columnwidth*0.5, 0)$);

    \draw[arrow] (conv-\b-2-1) to ($(joiner-\b-2-1.north) - (\columnwidth, 0)$);
    \draw[arrow,myround] (joiner-\b-2-1.south) to ($(joiner-\b-2-1.south) + (0, -\arrowbuf)$) to ($(joiner-\b-2-1.south) + (-\columnwidth, -\arrowbuf)$) to (conv-\b-2-2);
    \draw[arrow] (conv-\b-2-2) to ($(joiner-\b-1-1.north) + (-\columnwidth*0.5, 0)$);

    \draw[arrow] (conv-\b-1-1) to ($(joiner-\b-1-1.north) - (\columnwidth*1.5, 0)$);

    \node (exit) at (\one+\columnwidth*\columns/2, -\blockheight-1.2) {$f_4(z)$};

    \draw[arrow] (joiner-\b-1-1) to  (exit);
}

\begin{pgfonlayer}{background}
\filldraw [line width=4mm,join=round,black!10]
  (below-entry.south -| conv-1-4-1.east) rectangle (joiner-1-1-1.south -| conv-1-1-1.west);
\end{pgfonlayer}

\def\two{\one+\columnwidth*\columns + 2}
\def\betweenblocks{1.3}

\foreach \b in {1, ..., 5} {
    \node[rounded corners,black,fill=black!10,join=round,minimum width=2.0cm,minimum height=0.6cm] (block-\b) at (\two, -\blockheight/2 -\b*\betweenblocks + 3*\betweenblocks) {\scriptsize{\textsf{\textbf{Block \b}}}};
    \node[pool] (pool-\b) at (\two, -\blockheight/2 -\b*\betweenblocks + 2.5*\betweenblocks) {};
}

\node[dense] (dense) at ($(pool-5) + (0, -0.5)$) {};

\node[] (x) at (\two, 0.6) {$x$};
\node[] (y) at (\two,  -\blockheight-1.2) {$y$};
\draw[arrow] (x) to (block-1);
\draw[arrow] (block-1) to (pool-1);
\draw[arrow] (pool-1) to (block-2);
\draw[arrow] (block-2) to (pool-2);
\draw[arrow] (pool-2) to (block-3);
\draw[arrow] (block-3) to (pool-3);
\draw[arrow] (pool-3) to (block-4);
\draw[arrow] (block-4) to (pool-4);
\draw[arrow] (pool-4) to (block-5);
\draw[arrow] (block-5) to (pool-5);
\draw[arrow] (pool-5) to (dense);
\draw[arrow] (dense) to (y);

\draw[-,dashed,thick,black!30] (\two - 1.1, -\blockheight/2 + 0.4) -- (\two - 2.1, -\blockheight/2 + 3.3);
\draw[-,dashed,thick,black!30] (\two - 1.1, -\blockheight/2 - 0.4) -- (\two - 2.1, -\blockheight/2 - 3.3);

\def\three{0}
\def\lx{\three + 0.7 + 0.25}
\def\ly{-6}
\def\betweenlegends{0.5}

\def\leftcenter{\dividerx - 2.95}
\def\dividerx{\one-0.5}

\draw[-,dashed,thick,black] (\dividerx, 0.7) -- (\dividerx, -8.3);

\node[] (text-depth-expansion) at (\leftcenter+0.25, 0.5) {\textsf{\textbf{\footnotesize{Fractal Expansion Rule}}}};
\node[] (text-layer-guide) at (\leftcenter+0.25, \ly+0.7) {\textsf{\textbf{\footnotesize{Layer Key}}}};

\node[conv] (legend-conv) at (\lx, \ly) {};
\node[anchor=west] (legend-conv-text) at ($(legend-conv.east) + (0.1, 0)$) {\textsf{\textbf{\scriptsize{Convolution}}}};

\node[joiner] (legend-joiner) at (\lx, \ly-\betweenlegends) {};
\node[anchor=west] at ($(legend-joiner.east) + (0.1, 0)$) {\textsf{\textbf{\scriptsize{Join}}}};

\node[pool] (legend-pool) at (\lx, \ly-2*\betweenlegends) {};
\node[anchor=west] (legend-pool-text) at ($(legend-pool.east) + (0.1, 0)$) {\textsf{\textbf{\scriptsize{Pool}}}};

\node[dense] (legend-dense) at (\lx, \ly-3*\betweenlegends) {};
\node[anchor=west] (legend-dense-text) at ($(legend-dense.east) + (0.1, 0)$) {\textsf{\textbf{\scriptsize{Prediction}}}};

\def\buf{0.25}
\begin{pgfonlayer}{background}
\draw [line width=0.2mm,dashed,rounded corners]
  ($(legend-conv.north -| legend-conv.west) + (-\buf, \buf)$) rectangle ($(legend-dense.south -| legend-conv-text.east) + (\buf, -\buf)$);
\end{pgfonlayer}

\def\ex{\three}
\def\ey{0.0}
\def\exb{\three + 3.0}
\def\eheight{4.0}
\def\eradius{\columnwidth/2}

\node[] (ex-1) at (\ex, \ey) {$z$};
\node[func] (ef-1) at (\ex,\ey - 0.5*\eheight) {$f_{C}$};
\node[] (ey-1) at (\ex, \ey-\eheight) {$f_{C}(z)$};
\draw[arrow] (ex-1) to (ef-1);
\draw[arrow] (ef-1) to (ey-1);

\node[] (ex-2) at (\exb, \ey) {$z$};
\node[conv] (econv) at (\exb - \eradius,\ey - 0.45*\eheight) {};
\node[func] (ef-2-1) at (\exb + \eradius,\ey - 0.3*\eheight) {$f_{C}$};
\node[func] (ef-2-2) at (\exb + \eradius,\ey - 0.6*\eheight) {$f_{C}$};
\node[joiner3] (ejoin) at (\exb,\ey - 0.8*\eheight) {};
\node[] (ey-2) at (\exb, \ey-\eheight) {$f_{C+1}(z)$};

\draw[arrow,rounded corners] (ex-2) to ($(ex-2) + (0, -0.3)$) to ($(ex-2) + (\eradius, -0.3)$) to (ef-2-1);
\draw[arrow,rounded corners] (ex-2) to ($(ex-2) + (0, -0.3)$) to ($(ex-2) + (-\eradius, -0.3)$) to (econv);
\draw[arrow] (ef-2-1) to (ef-2-2);
\draw[arrow] (econv) to ($(ejoin.north) + (-\eradius, 0)$);
\draw[arrow] (ef-2-2) to ($(ejoin.north) + (+\eradius, 0)$);
\draw[arrow] (ejoin.south) to (ey-2);

\node[single arrow,rotate=0,fill=black!20] at (\exb - 1.85, \ey - 0.5*\eheight) {\phantom{hej}};

\end{tikzpicture}

%% file: include/sec2-related.tex
Deepening feed-forward neural networks has generally returned dividends in
performance.  A striking example within the computer vision community is the
improvement on the ImageNet~\citep{deng2009imagenet} classification task when
transitioning from AlexNet~\citep{alexNet12} to VGG~\citep{vgg16} to
GoogLeNet~\citep{szegedy2015inception} to {\resnet}~\citep{he2015deep}.
Unfortunately, greater depth also makes training more challenging, at least
when employing a first-order optimization method with randomly initialized
layers.  As the network grows deeper and more non-linear, the linear
approximation of a gradient step becomes increasingly inappropriate.  Desire
to overcome these difficulties drives research on both optimization techniques
and network architectures.

On the optimization side, much recent work yields improvements.  To prevent
vanishing gradients, ReLU activation functions now widely replace sigmoid and
tanh units~\citep{nair2010rectified}.  This subject remains an area of active
inquiry, with various tweaks on ReLUs, \eg~PReLUs~\citep{he2015prelu}, and
ELUs~\citep{elu}.  Even with ReLUs, employing batch normalization~\citep{
batchnorm} speeds training by reducing internal covariate shift.  Good
initialization can also ameliorate this problem~\citep{glorot2010understanding,
mishkin2015all}.  Path-SGD~\citep{pathsgd} offers an alternative normalization
scheme.  Progress in optimization is somewhat orthogonal to our architectural
focus, with the expectation that advances in either are ripe for combination.

Notable ideas in architecture reach back to skip connections, the earliest
example of a nontrivial routing pattern within a neural network.  Recent work
further elaborates upon them~\citep{MYP:ACCV:2014,HAGM:CVPR:2015}.  Highway
networks~\citep{srivastava2015highway} and {\resnet}~\citep{he2015deep,
he2016identity} offer additional twists in the form of parameterized
pass-through and gating.  In work subsequent to our own, \cite{densenet}
investigate a {\resnet} variant with explicit skip connections.  These methods
share distinction as the only other designs demonstrated to scale to hundreds
of layers and beyond.  {\Resnet}'s building block uses the identity map as an
anchor point and explicitly parameterizes an additive correction term (the
residual).  Identity initialization also appears in the context of recurrent
networks~\citep{le2015simple}.  A tendency of {\resnet} and highway networks to
fall-back to the identity map may make their effective depth much smaller than
their nominal depth.

Some prior results hint at what we experimentally demonstrate in
Section~\ref{sec:results}.  Namely, reduction of effective depth is key to
training extremely deep networks; residuals are incidental.
\cite{huang2016stochasticdepth} provide one clue in their work on stochastic
depth: randomly dropping layers from {\resnet} during training, thereby
shrinking network depth by a constant factor, provides additional performance
benefit.  We build upon this intuition through {\droppath}, which shrinks depth
much more drastically.

The success of deep supervision~\citep{lee2014deeply} provides another clue
that effective depth is crucial.  Here, an auxiliary loss, forked off mid-level
layers, introduces a shorter path during backpropagation.  The layer at the
fork receives two gradients, originating from the main loss and the auxiliary
loss, that are added together.  Deep supervision is now common, being adopted,
for example, by GoogLeNet~\citep{szegedy2015inception}.  However, irrelevance
of the auxiliary loss at test time introduces the drawback of having a
discrepancy between the actual objective and that used for training.

Exploration of the student-teacher paradigm~\citep{ba2014dodeep} illuminates
the potential for interplay between networks of different depth.  In the model
compression scenario, a deeper network (previously trained) guides and improves
the learning of a shallower and faster student network~\citep{ba2014dodeep,
urban2016dodeepsfollowup}.  This is accomplished by feeding unlabeled data
through the teacher and having the student mimic the teacher's soft output
predictions.  FitNets~\citep{romero2014fitnets} explicitly couple students and
teachers, forcing mimic behavior across several intermediate points in the
network.  Our fractal networks capture yet another alternative, in the form of
implicit coupling, with the potential for bidirectional information flow
between shallow and deep subnetworks.

Widening networks, by using larger modules in place of individual layers, has
also produced performance gains.  For example, an Inception module~\citep{
szegedy2015inception} concatenates results of convolutional layers of different
receptive field size.  Stacking these modules forms the GoogLeNet architecture.
\cite{liao2015competitive} employ a variant with maxout in place of
concatenation.  Figure~\ref{fig:fractalnet} makes apparent our connection with
such work.  As a fractal network deepens, it also widens.  Moreover, note that
stacking two 2D convolutional layers with the same spatial receptive field
(\eg~$3 \times 3$) achieves a larger ($5 \times 5$) receptive field.  A
horizontal cross-section of a fractal network is reminiscent of an Inception
module, except with additional joins due to recursive structure.

%% file: include/sec3-method.tex
We begin with a more formal presentation of the ideas sketched in
Figure~\ref{fig:fractalnet}.  Convolutional neural networks serve as our
running example and, in the subsequent section, our experimental platform.
However, it is worth emphasizing that our framework is more general.  In
principle, convolutional layers in Figure~\ref{fig:fractalnet} could be
replaced by a different layer type, or even a custom-designed module or
subnetwork, in order to generate other fractal architectures.

Let $C$ denote the index of the truncated fractal $f_C(\cdot)$.  Our network's
structure, connections and layer types, is defined by $f_C(\cdot)$.  A network
consisting of a single convolutional layer is the base case:
\begin{equation}
   f_1(z) = \mathrm{conv}(z)
   \label{eq:frac-base}
\end{equation}
We define successive fractals recursively:
\begin{equation}
    f_{C+1}(z) =
      \left[(f_C \circ f_C)(z)\right] \oplus
      \left[\mathrm{conv}(z)\right]
   \label{eq:frac-gen}
\end{equation}
where $\circ$ denotes composition and $\oplus$ a join operation.  When drawn
in the style of Figure~\ref{fig:fractalnet}, $C$ corresponds to the number of
columns, or width, of network $f_C(\cdot)$.  Depth, defined to be the number of
$\mathrm{conv}$ layers on the longest path between input and output, scales as
$2^{C-1}$.  Convolutional networks for classification typically intersperse
pooling layers.  We achieve the same by using $f_C(\cdot)$ as a building block
and stacking it with subsequent pooling layers $B$ times, yielding total depth
$B \cdot 2^{C-1}$.

The join operation $\oplus$ merges two feature blobs into one.  Here, a blob
is the result of a $\mathrm{conv}$ layer: a tensor holding activations for a
fixed number of channels over a spatial domain.  The channel count corresponds
to the size of the filter set in the preceding $\mathrm{conv}$ layer.  As the
fractal is expanded, we collapse neighboring joins into a single
$\mathrm{join}$ layer which spans multiple columns, as shown on the right side
of Figure~\ref{fig:fractalnet}.  The $\mathrm{join}$ layer merges all of its
input feature blobs into a single output blob.

Several choices seem reasonable for the action of a $\mathrm{join}$ layer,
including concatenation and addition.  We instantiate each $\mathrm{join}$
to compute the element-wise mean of its inputs.  This is appropriate for
convolutional networks in which channel count is set the same for all
$\mathrm{conv}$ layers within a fractal block.  Averaging might appear similar
to {\resnet}'s addition operation, but there are critical differences:
\begin{itemize}
   \item{
      {\Resnet} makes clear distinction between pass-through and residual
      signals.  In {\fracnet}, no signal is privileged.  Every input to a
      $\mathrm{join}$ layer is the output of an immediately preceding
      $\mathrm{conv}$ layer.  The network structure alone cannot identify any
      as being primary.
   }
   \item{
      {\Droppath} regularization, as described next in
      Section~\ref{sec:droppath}, forces each input to a $\mathrm{join}$ to
      be individually reliable.  This reduces the reward for even implicitly
      learning to allocate part of one signal to act as a residual for
      another.
   }
   \item{
      Experiments show that we can extract high-performance subnetworks
      consisting of a single column (Section~\ref{sec:evaluation}).  Such a
      subnetwork is effectively devoid of joins, as only a single path is
      active throughout.  They produce no signal to which a residual could be
      added.
   }
\end{itemize}
Together, these properties ensure that $\mathrm{join}$ layers are not an
alternative method of residual learning.

\subsection{Regularization via {\Droppath}}
\label{sec:droppath}

\begin{figure}[t]
   \begin{center}
        \begingroup
        \input{#include/fig-droppath}%
        \endgroup
    
   \end{center}
   \vspace{-0.02\linewidth}
   \caption{
      \textbf{{\Droppath}.}
         A fractal network block functions with some connections between layers
         disabled, provided some path from input to output is still available.
         {\Droppath} guarantees at least one such path, while sampling a
         subnetwork with many other paths disabled.  During training,
         presenting a different active subnetwork to each mini-batch prevents
         co-adaptation of parallel paths.  A global sampling strategy
         returns a single column as a subnetwork.  Alternating it with local
         sampling encourages the development of individual columns as
         performant stand-alone subnetworks.
   }
   \label{fig:droppath}
\end{figure}
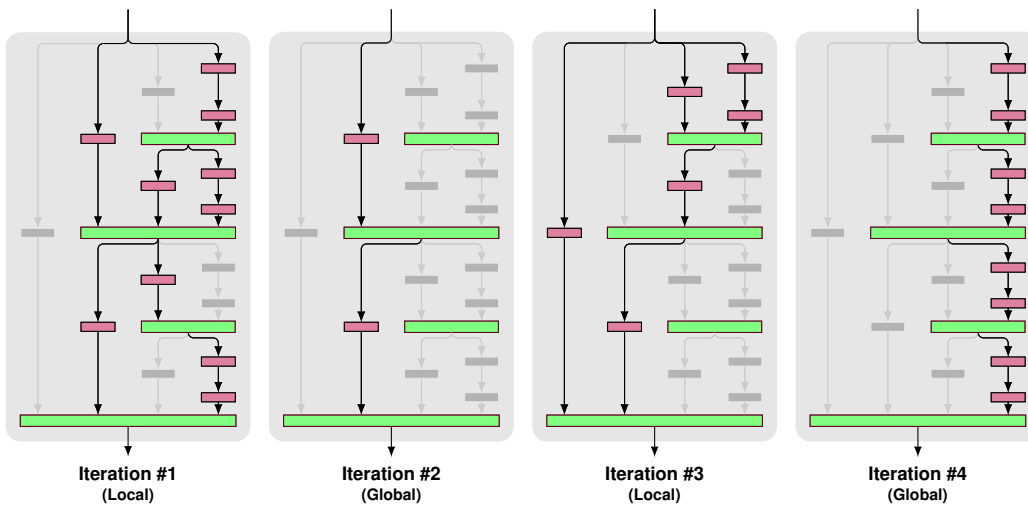

{\Dropout}~\citep{dropout} and {\dropconn}~\citep{dropconnect} modify
interactions between sequential network layers in order to discourage
co-adaptation.  Since fractal networks contain additional macro-scale
structure, we propose to complement these techniques with an analogous
coarse-scale regularization scheme.

Figure~\ref{fig:droppath} illustrates {\droppath}.  Just as {\dropout} prevents
co-adaptation of activations, {\droppath} prevents co-adaptation of parallel
paths by randomly dropping operands of the $\mathrm{join}$ layers.  This
discourages the network from using one input path as an anchor and another as a
corrective term (a configuration that, if not prevented, is prone to
overfitting).  We consider two sampling strategies:
\begin{itemize}
   \item{
      \textbf{Local}: a $\mathrm{join}$ drops each input with fixed
      probability, but we make sure at least one survives.
   }
   \item{
      \textbf{Global}: a single path is selected for the entire network.
      We restrict this path to be a single column, thereby promoting
      individual columns as independently strong predictors.
   }
\end{itemize}
As with {\dropout}, signals may need appropriate rescaling.  With element-wise
means, this is trivial; each $\mathrm{join}$ computes the mean of only its
active inputs.

In experiments, we train with {\dropout} and a mixture model of $50\%$ local
and $50\%$ global sampling for {\droppath}.  We sample a new subnetwork each
mini-batch.  With sufficient memory, we can simultaneously evaluate one local
sample and all global samples for each mini-batch by keeping separate
networks and tying them together via weight sharing.

While fractal connectivity permits the use of paths of any length, global
{\droppath} forces the use of many paths whose lengths differ by orders of
magnitude (powers of $2$).  The subnetworks sampled by {\droppath} thus exhibit
large structural diversity.  This property stands in contrast to stochastic
depth regularization of {\resnet}, which, by virtue of using a fixed drop
probability for each layer in a chain, samples subnetworks with a concentrated
depth distribution~\citep{huang2016stochasticdepth}.

Global {\droppath} serves not only as a regularizer, but also as a diagnostic
tool.  Monitoring performance of individual columns provides insight into both
the network and training mechanisms, as Section~\ref{sec:introspection}
discusses in more detail.  Individually strong columns of various depths also
give users choices in the trade-off between speed (shallow) and accuracy
(deep).

\subsection{Data Augmentation}
\label{sec:data-aug}

Data augmentation can reduce the need for regularization.  {\Resnet}
demonstrates this, achieving 27.22\% error rate on CIFAR-100 with augmentation
compared to 44.76\% without~\citep{huang2016stochasticdepth}.  While
augmentation benefits fractal networks, we show that {\droppath} provides
highly effective regularization, allowing them to achieve competitive results
even without data augmentation.

\subsection{Implementation Details}
\label{sec:implementation}

We implement {\fracnet} using Caffe~\citep{caffe14}.  Purely for convenience,
we flip the order of pool and join layers at the end of a block in
Figure~\ref{fig:fractalnet}.  We pool individual columns immediately before
the joins spanning all columns, rather than pooling once immediately after
them.

We train fractal networks using stochastic gradient descent with momentum.  As
now standard, we employ batch normalization together with each $\mathrm{conv}$
layer (convolution, batch norm, then ReLU).

%% file: include/fig-droppath.tex
\begin{tikzpicture}[->,>=latex,auto,node distance=3cm,transform shape,
 thick,main node/.style={circle,draw,font=\sffamily\Large\bfseries}]
\def\blockheight{5.0}
\def\layerwidth{0.3}
\def\layerheight{0.04}
\def\columnwidth{0.8}
\def\iterwidth{3.5}
\def\one{0}
\def\two{\iterwidth}
\def\three{2*\iterwidth}
\def\four{3*\iterwidth}
\def\extrabufy{0.15}

\def\blocks{1}
\def\columns{4}

\def\arrowbuf{0.07}
\tikzstyle{noconv}=[rectangle,draw=black!10,thick,fill=black!30,thin,minimum width=1.5cm,minimum height=0.4cm,scale=0.3]
\tikzstyle{conv}=[rectangle,draw,thick,fill=purple!50,thin,minimum width=1.5cm,minimum height=0.4cm,scale=0.3]
\tikzstyle{pool}=[rectangle,draw,thick,fill=yellow!50,thin,minimum width=1.5cm,minimum height=0.4cm,scale=0.3]
\tikzstyle{dense}=[rectangle,draw,thick,fill=blue!50,thin,minimum width=1.5cm,minimum height=0.4cm,scale=0.3]
\tikzstyle{func}=[rectangle,draw,thick,fill=gray!50,thin,scale=1.0]
\tikzstyle{joiner}=[rectangle,thick,draw,fill=green!50,draw=black!50!purple,thin,minimum width=1.5cm,minimum height=0.5cm,scale=0.3]
\tikzstyle{joiner1}=[rectangle,thick,draw,fill=green!50,draw=black!50!purple,thin,minimum width=9.55cm,minimum height=0.5cm,scale=0.3]
\tikzstyle{joiner2}=[rectangle,thick,draw,fill=green!50,draw=black!50!purple,thin,minimum width=6.85cm,minimum height=0.5cm,scale=0.3]
\tikzstyle{joiner3}=[rectangle,thick,draw,fill=green!50,draw=black!50!purple,thin,minimum width=4.15cm,minimum height=0.5cm,scale=0.3]
\tikzstyle{myround}=[rounded corners=0.07cm]
\tikzstyle{arrow}=[->,thin]
\tikzstyle{noarrow}=[->,thin,black!20]

\foreach \b in {1, \blocks} {
    \node[joiner1] (joiner-\b-1-1) at (\one+\columnwidth*\columns/2, -\b*\blockheight) {};
    \foreach \j in {1, ..., 1} {
        \node[joiner2] (joiner-\b-2-\j) at (\one+\columnwidth*2.5, -\b*\blockheight +\blockheight -\blockheight/2 - \j*\blockheight/1 + \blockheight/1) {};
    }
    \foreach \j in {1, ..., 2} {
        \node[joiner3] (joiner-\b-3-\j) at (\one+\columnwidth*3, -\b*\blockheight +\blockheight -\blockheight/4 - \j*\blockheight/2 + \blockheight/2) {};
    }
    \foreach \c/\layers in {1/1, 2/2, 3/4, 4/8} {
        \def\x{\one+\c*\columnwidth - \columnwidth/2}
        \def\y{\blockheight-\b*\blockheight}
        \def\convradius{\blockheight/\layers}
        \foreach \n in {1, ..., \layers} {
            \ifthenelse{\c=4 \AND \n=3}{
                \def\yn{\y - \n*\convradius + 0.5*\convradius - \extrabufy}
            }{
                \ifthenelse{\c=4 \AND \n=5}{
                    \def\yn{\y - \n*\convradius + 0.5*\convradius - \extrabufy}
                }{
                    \ifthenelse{\c=4 \AND \n=7}{
                        \def\yn{\y - \n*\convradius + 0.5*\convradius - \extrabufy}
                    }{
                        \def\yn{\y - \n*\convradius + 0.5*\convradius}
                    }
                }
            }
            \def\xn{\x}
            \ifthenelse{\c=1 \AND \n=1 \OR \c=3 \AND \n=1}{
                \node[noconv] (conv-\b-\c-\n) at (\xn, \yn) {};
            }{
                \ifthenelse{\c=4 \AND \n=5}{
                    \node[noconv] (conv-\b-\c-\n) at (\xn, \yn) {};
                }{
                    \ifthenelse{\c=4 \AND \n=6}{
                        \node[noconv] (conv-\b-\c-\n) at (\xn, \yn) {};
                    }{
                        \ifthenelse{\c=3 \AND \n=4}{
                            \node[noconv] (conv-\b-\c-\n) at (\xn, \yn) {};
                        }{
                            \node[conv] (conv-\b-\c-\n) at (\xn, \yn) {};
                        }
                    }
                }
            }
        }
    }

    \node (entry) at (\one+\columnwidth*\columns/2, 0.6) {};
    \node (below-entry) at (\one+\columnwidth*\columns/2, 0.1) {};

    \draw[noarrow,myround] (entry.south) to ($(entry.south) + (0, -0.45)$) to ($(entry.south) + (+\columnwidth*0.5, -0.45)$) to  (conv-\b-3-1);
    \draw[noarrow,myround] (entry.south) to ($(entry.south) + (0, -0.45)$) to ($(entry.south) + (-\columnwidth*1.5, -0.45)$) to  (conv-\b-1-1);
    \draw[arrow,myround] (entry.south) to ($(entry.south) + (0, -0.45)$) to ($(entry.south) + (+\columnwidth*1.5, -0.45)$) to  (conv-\b-4-1);
    \draw[arrow,myround] (entry.south) to ($(entry.south) + (0, -0.45)$) to ($(entry.south) + (-\columnwidth*0.5, -0.45)$) to  (conv-\b-2-1);

    \draw[noarrow,myround] (joiner-\b-2-1.south) to ($(joiner-\b-2-1.south) + (0, -\arrowbuf)$) to ($(joiner-\b-2-1.south) + (\columnwidth, -\arrowbuf)$) to (conv-\b-4-5);
    \draw[noarrow] (conv-\b-4-5) to (conv-\b-4-6);
    \draw[noarrow] (conv-\b-4-6) to ($(joiner-\b-3-2.north) + (\columnwidth/2, 0)$);
    \draw[noarrow,myround] (joiner-\b-3-2.south) to ($(joiner-\b-3-2.south) + (0, -\arrowbuf)$) to ($(joiner-\b-3-2.south) + (-\columnwidth/2, -\arrowbuf)$) to (conv-\b-3-4);
    \draw[noarrow] (conv-\b-3-4) to ($(joiner-\b-1-1.north) + (\columnwidth*0.5, 0)$);
    \draw[noarrow] (conv-\b-1-1) to ($(joiner-\b-1-1.north) - (\columnwidth*1.5, 0)$);
    \draw[arrow] (conv-\b-4-1) to (conv-\b-4-2);
    \draw[arrow] (conv-\b-4-2) to ($(joiner-\b-3-1.north) + (\columnwidth/2, 0)$);
    \draw[arrow,myround] (joiner-\b-3-1.south) to ($(joiner-\b-3-1.south) + (0, -\arrowbuf)$) to ($(joiner-\b-3-1.south) + (\columnwidth/2, -\arrowbuf)$) to (conv-\b-4-3);
    \draw[arrow] (conv-\b-4-3) to (conv-\b-4-4);
    \draw[arrow] (conv-\b-4-4) to ($(joiner-\b-2-1.north) + (\columnwidth, 0)$);
    \draw[arrow,myround] (joiner-\b-3-2.south) to ($(joiner-\b-3-2.south) + (0, -\arrowbuf)$) to ($(joiner-\b-3-2.south) + (\columnwidth/2, -\arrowbuf)$) to (conv-\b-4-7);
    \draw[arrow] (conv-\b-4-7) to (conv-\b-4-8);
    \draw[arrow] (conv-\b-4-8) to ($(joiner-\b-1-1.north) + (\columnwidth*1.5, 0)$);

    \draw[noarrow] (conv-\b-3-1) to ($(joiner-\b-3-1.north) - (\columnwidth/2, 0)$);
    \draw[arrow,myround] (joiner-\b-3-1.south) to ($(joiner-\b-3-1.south) + (0, -\arrowbuf)$)  to ($(joiner-\b-3-1.south) + (-\columnwidth/2, -\arrowbuf)$)  to (conv-\b-3-2);
    \draw[arrow] (conv-\b-3-2) to (joiner-\b-2-1);
    \draw[arrow] (joiner-\b-2-1) to (conv-\b-3-3);
    \draw[arrow] (conv-\b-3-3) to ($(joiner-\b-3-2.north) - (\columnwidth/2, 0)$);

    \draw[arrow] (conv-\b-2-1) to ($(joiner-\b-2-1.north) - (\columnwidth, 0)$);
    \draw[arrow,myround] (joiner-\b-2-1.south) to ($(joiner-\b-2-1.south) + (0, -\arrowbuf)$) to ($(joiner-\b-2-1.south) + (-\columnwidth, -\arrowbuf)$) to (conv-\b-2-2);
    \draw[arrow] (conv-\b-2-2) to ($(joiner-\b-1-1.north) + (-\columnwidth*0.5, 0)$);

}

\node (exit) at (\one+\columnwidth*\columns/2, -\blocks*\blockheight-0.7) {\textbf{\textsf{\scriptsize{Iteration \#1}}}};
\node (exit-south) at (\one+\columnwidth*\columns/2, -\blocks*\blockheight-1.0) {\textbf{\textsf{\tiny{(Local)}}}};
\draw[arrow] (joiner-1-1-1) to  (exit);

\begin{pgfonlayer}{background}
\filldraw [line width=4mm,join=round,black!10]
  (below-entry.south -| conv-1-4-1.east) rectangle (joiner-1-1-1.south -| conv-1-1-1.west);
\end{pgfonlayer}

\foreach \b in {1, \blocks} {
    \node[joiner1] (joiner-\b-1-1) at (\two+\columnwidth*\columns/2, -\b*\blockheight) {};
    \foreach \j in {1, ..., 1} {
        \node[joiner2] (joiner-\b-2-\j) at (\two+\columnwidth*2.5, -\b*\blockheight +\blockheight -\blockheight/2 - \j*\blockheight/1 + \blockheight/1) {};
    }
    \foreach \j in {1, ..., 2} {
        \node[joiner3] (joiner-\b-3-\j) at (\two+\columnwidth*3, -\b*\blockheight +\blockheight -\blockheight/4 - \j*\blockheight/2 + \blockheight/2) {};
    }
    \foreach \c/\layers in {1/1, 2/2, 3/4, 4/8} {
        \def\x{\two+\c*\columnwidth - \columnwidth/2}
        \def\y{\blockheight-\b*\blockheight}
        \def\convradius{\blockheight/\layers}
        \foreach \n in {1, ..., \layers} {
            \ifthenelse{\c=4 \AND \n=3}{
                \def\yn{\y - \n*\convradius + 0.5*\convradius - \extrabufy}
            }{
                \ifthenelse{\c=4 \AND \n=5}{
                    \def\yn{\y - \n*\convradius + 0.5*\convradius - \extrabufy}
                }{
                    \ifthenelse{\c=4 \AND \n=7}{
                        \def\yn{\y - \n*\convradius + 0.5*\convradius - \extrabufy}
                    }{
                        \def\yn{\y - \n*\convradius + 0.5*\convradius}
                    }
                }
            }
            \def\xn{\x}
            \ifthenelse{\c=2}{
                \node[conv] (conv-\b-\c-\n) at (\xn, \yn) {};
            }{
                \node[noconv] (conv-\b-\c-\n) at (\xn, \yn) {};
            }
        }
    }

    \node (entry) at (\two+\columnwidth*\columns/2, 0.6) {};
    \node (below-entry) at (\two+\columnwidth*\columns/2, 0.1) {};

    \draw[noarrow,myround] (entry.south) to ($(entry.south) + (0, -0.45)$) to ($(entry.south) + (+\columnwidth*0.5, -0.45)$) to  (conv-\b-3-1);
    \draw[noarrow,myround] (entry.south) to ($(entry.south) + (0, -0.45)$) to ($(entry.south) + (-\columnwidth*1.5, -0.45)$) to  (conv-\b-1-1);
    \draw[noarrow,myround] (entry.south) to ($(entry.south) + (0, -0.45)$) to ($(entry.south) + (+\columnwidth*1.5, -0.45)$) to  (conv-\b-4-1);
    \draw[arrow,myround] (entry.south) to ($(entry.south) + (0, -0.45)$) to ($(entry.south) + (-\columnwidth*0.5, -0.45)$) to  (conv-\b-2-1);

    \draw[noarrow] (conv-\b-4-1) to (conv-\b-4-2);
    \draw[noarrow] (conv-\b-4-2) to ($(joiner-\b-3-1.north) + (\columnwidth/2, 0)$);
    \draw[noarrow,myround] (joiner-\b-3-1.south) to ($(joiner-\b-3-1.south) + (0, -\arrowbuf)$) to ($(joiner-\b-3-1.south) + (\columnwidth/2, -\arrowbuf)$) to (conv-\b-4-3);
    \draw[noarrow] (conv-\b-4-3) to (conv-\b-4-4);
    \draw[noarrow] (conv-\b-4-4) to ($(joiner-\b-2-1.north) + (\columnwidth, 0)$);
    \draw[noarrow,myround] (joiner-\b-2-1.south) to ($(joiner-\b-2-1.south) + (0, -\arrowbuf)$) to ($(joiner-\b-2-1.south) + (\columnwidth, -\arrowbuf)$) to (conv-\b-4-5);
    \draw[noarrow] (conv-\b-4-5) to (conv-\b-4-6);
    \draw[noarrow] (conv-\b-4-6) to ($(joiner-\b-3-2.north) + (\columnwidth/2, 0)$);
    \draw[noarrow,myround] (joiner-\b-3-2.south) to ($(joiner-\b-3-2.south) + (0, -\arrowbuf)$) to ($(joiner-\b-3-2.south) + (\columnwidth/2, -\arrowbuf)$) to (conv-\b-4-7);
    \draw[noarrow] (conv-\b-4-7) to (conv-\b-4-8);
    \draw[noarrow] (conv-\b-4-8) to ($(joiner-\b-1-1.north) + (\columnwidth*1.5, 0)$);

    \draw[noarrow] (conv-\b-3-1) to ($(joiner-\b-3-1.north) - (\columnwidth/2, 0)$);
    \draw[noarrow,myround] (joiner-\b-3-1.south) to ($(joiner-\b-3-1.south) + (0, -\arrowbuf)$)  to ($(joiner-\b-3-1.south) + (-\columnwidth/2, -\arrowbuf)$)  to (conv-\b-3-2);
    \draw[noarrow] (conv-\b-3-2) to (joiner-\b-2-1);
    \draw[noarrow] (joiner-\b-2-1) to (conv-\b-3-3);
    \draw[noarrow] (conv-\b-3-3) to ($(joiner-\b-3-2.north) - (\columnwidth/2, 0)$);
    \draw[noarrow,myround] (joiner-\b-3-2.south) to ($(joiner-\b-3-2.south) + (0, -\arrowbuf)$) to ($(joiner-\b-3-2.south) + (-\columnwidth/2, -\arrowbuf)$) to (conv-\b-3-4);
    \draw[noarrow] (conv-\b-3-4) to ($(joiner-\b-1-1.north) + (\columnwidth*0.5, 0)$);

    \draw[noarrow] (conv-\b-1-1) to ($(joiner-\b-1-1.north) - (\columnwidth*1.5, 0)$);

    \draw[arrow] (conv-\b-2-1) to ($(joiner-\b-2-1.north) - (\columnwidth, 0)$);
    \draw[arrow,myround] (joiner-\b-2-1.south) to ($(joiner-\b-2-1.south) + (0, -\arrowbuf)$) to ($(joiner-\b-2-1.south) + (-\columnwidth, -\arrowbuf)$) to (conv-\b-2-2);
    \draw[arrow] (conv-\b-2-2) to ($(joiner-\b-1-1.north) + (-\columnwidth*0.5, 0)$);

}

\node (exit) at (\two+\columnwidth*\columns/2, -\blocks*\blockheight-0.7) {\textbf{\textsf{\scriptsize{Iteration \#2}}}};
\node (exit-south) at (\two+\columnwidth*\columns/2, -\blocks*\blockheight-1.0) {\textbf{\textsf{\tiny{(Global)}}}};
\draw[arrow] (joiner-1-1-1) to  (exit);

\begin{pgfonlayer}{background}
\filldraw [line width=4mm,join=round,black!10]
  (below-entry.south -| conv-1-4-1.east) rectangle (joiner-1-1-1.south -| conv-1-1-1.west);
\end{pgfonlayer}

\foreach \b in {1, \blocks} {
    \node[joiner1] (joiner-\b-1-1) at (\three+\columnwidth*\columns/2, -\b*\blockheight) {};
    \foreach \j in {1, ..., 1} {
        \node[joiner2] (joiner-\b-2-\j) at (\three+\columnwidth*2.5, -\b*\blockheight +\blockheight -\blockheight/2 - \j*\blockheight/1 + \blockheight/1) {};
    }
    \foreach \j in {1, ..., 2} {
        \node[joiner3] (joiner-\b-3-\j) at (\three+\columnwidth*3, -\b*\blockheight +\blockheight -\blockheight/4 - \j*\blockheight/2 + \blockheight/2) {};
    }
    \foreach \c/\layers in {1/1, 2/2, 3/4, 4/8} {
        \def\x{\three+\c*\columnwidth - \columnwidth/2}
        \def\y{\blockheight-\b*\blockheight}
        \def\convradius{\blockheight/\layers}
        \foreach \n in {1, ..., \layers} {
            \ifthenelse{\c=4 \AND \n=3}{
                \def\yn{\y - \n*\convradius + 0.5*\convradius - \extrabufy}
            }{
                \ifthenelse{\c=4 \AND \n=5}{
                    \def\yn{\y - \n*\convradius + 0.5*\convradius - \extrabufy}
                }{
                    \ifthenelse{\c=4 \AND \n=7}{
                        \def\yn{\y - \n*\convradius + 0.5*\convradius - \extrabufy}
                    }{
                        \def\yn{\y - \n*\convradius + 0.5*\convradius}
                    }
                }
            }
            \def\xn{\x}
            \ifthenelse{\c=2 \AND \n=1}{
                \node[noconv] (conv-\b-\c-\n) at (\xn, \yn) {};
            }{
                \ifthenelse{\c=3 \AND \n=3}{
                    \node[noconv] (conv-\b-\c-\n) at (\xn, \yn) {};
                }{
                    \ifthenelse{\c=3 \AND \n=4}{
                        \node[noconv] (conv-\b-\c-\n) at (\xn, \yn) {};
                    }{

                        \ifthenelse{\c=4}{
                            \ifthenelse{\n=1 \OR \n=2}{
                                \node[conv] (conv-\b-\c-\n) at (\xn, \yn) {};
                            }{
                                \node[noconv] (conv-\b-\c-\n) at (\xn, \yn) {};
                            }
                        }{
                            \node[conv] (conv-\b-\c-\n) at (\xn, \yn) {};
                        }
                    }
                }
            }
        }
    }

    \node (entry) at (\three+\columnwidth*\columns/2, 0.6) {};
    \node (below-entry) at (\three+\columnwidth*\columns/2, 0.1) {};

    \draw[noarrow,myround] (entry.south) to ($(entry.south) + (0, -0.45)$) to ($(entry.south) + (-\columnwidth*0.5, -0.45)$) to  (conv-\b-2-1);
    \draw[arrow,myround] (entry.south) to ($(entry.south) + (0, -0.45)$) to ($(entry.south) + (+\columnwidth*0.5, -0.45)$) to  (conv-\b-3-1);
    \draw[arrow,myround] (entry.south) to ($(entry.south) + (0, -0.45)$) to ($(entry.south) + (-\columnwidth*1.5, -0.45)$) to  (conv-\b-1-1);
    \draw[arrow,myround] (entry.south) to ($(entry.south) + (0, -0.45)$) to ($(entry.south) + (+\columnwidth*1.5, -0.45)$) to  (conv-\b-4-1);

    \draw[arrow] (conv-\b-4-1) to (conv-\b-4-2);
    \draw[arrow] (conv-\b-4-2) to ($(joiner-\b-3-1.north) + (\columnwidth/2, 0)$);
    \draw[noarrow,myround] (joiner-\b-3-1.south) to ($(joiner-\b-3-1.south) + (0, -\arrowbuf)$) to ($(joiner-\b-3-1.south) + (\columnwidth/2, -\arrowbuf)$) to (conv-\b-4-3);
    \draw[noarrow] (conv-\b-4-3) to (conv-\b-4-4);
    \draw[noarrow] (conv-\b-4-4) to ($(joiner-\b-2-1.north) + (\columnwidth, 0)$);
    \draw[noarrow,myround] (joiner-\b-2-1.south) to ($(joiner-\b-2-1.south) + (0, -\arrowbuf)$) to ($(joiner-\b-2-1.south) + (\columnwidth, -\arrowbuf)$) to (conv-\b-4-5);
    \draw[noarrow] (conv-\b-4-5) to (conv-\b-4-6);
    \draw[noarrow] (conv-\b-4-6) to ($(joiner-\b-3-2.north) + (\columnwidth/2, 0)$);
    \draw[noarrow,myround] (joiner-\b-3-2.south) to ($(joiner-\b-3-2.south) + (0, -\arrowbuf)$) to ($(joiner-\b-3-2.south) + (\columnwidth/2, -\arrowbuf)$) to (conv-\b-4-7);
    \draw[noarrow] (conv-\b-4-7) to (conv-\b-4-8);
    \draw[noarrow] (conv-\b-4-8) to ($(joiner-\b-1-1.north) + (\columnwidth*1.5, 0)$);

    \draw[noarrow] (joiner-\b-2-1) to (conv-\b-3-3);
    \draw[noarrow] (conv-\b-3-3) to ($(joiner-\b-3-2.north) - (\columnwidth/2, 0)$);
    \draw[noarrow,myround] (joiner-\b-3-2.south) to ($(joiner-\b-3-2.south) + (0, -\arrowbuf)$) to ($(joiner-\b-3-2.south) + (-\columnwidth/2, -\arrowbuf)$) to (conv-\b-3-4);
    \draw[noarrow] (conv-\b-3-4) to ($(joiner-\b-1-1.north) + (\columnwidth*0.5, 0)$);

    \draw[noarrow] (conv-\b-2-1) to ($(joiner-\b-2-1.north) - (\columnwidth, 0)$);

    \draw[arrow] (conv-\b-3-1) to ($(joiner-\b-3-1.north) - (\columnwidth/2, 0)$);
    \draw[arrow,myround] (joiner-\b-3-1.south) to ($(joiner-\b-3-1.south) + (0, -\arrowbuf)$)  to ($(joiner-\b-3-1.south) + (-\columnwidth/2, -\arrowbuf)$)  to (conv-\b-3-2);
    \draw[arrow] (conv-\b-3-2) to (joiner-\b-2-1);
    \draw[arrow,myround] (joiner-\b-2-1.south) to ($(joiner-\b-2-1.south) + (0, -\arrowbuf)$) to ($(joiner-\b-2-1.south) + (-\columnwidth, -\arrowbuf)$) to (conv-\b-2-2);
    \draw[arrow] (conv-\b-2-2) to ($(joiner-\b-1-1.north) + (-\columnwidth*0.5, 0)$);

    \draw[arrow] (conv-\b-1-1) to ($(joiner-\b-1-1.north) - (\columnwidth*1.5, 0)$);

}

\node (exit) at (\three+\columnwidth*\columns/2, -\blocks*\blockheight-0.7) {\textbf{\textsf{\scriptsize{Iteration \#3}}}};
\node (exit-south) at (\three+\columnwidth*\columns/2, -\blocks*\blockheight-1.0) {\textbf{\textsf{\tiny{(Local)}}}};
\draw[arrow] (joiner-1-1-1) to  (exit);

\begin{pgfonlayer}{background}
\filldraw [line width=4mm,join=round,black!10]
  (below-entry.south -| conv-1-4-1.east) rectangle (joiner-1-1-1.south -| conv-1-1-1.west);
\end{pgfonlayer}

\foreach \b in {1, \blocks} {
    \node[joiner1] (joiner-\b-1-1) at (\four+\columnwidth*\columns/2, -\b*\blockheight) {};
    \foreach \j in {1, ..., 1} {
        \node[joiner2] (joiner-\b-2-\j) at (\four+\columnwidth*2.5, -\b*\blockheight +\blockheight -\blockheight/2 - \j*\blockheight/1 + \blockheight/1) {};
    }
    \foreach \j in {1, ..., 2} {
        \node[joiner3] (joiner-\b-3-\j) at (\four+\columnwidth*3, -\b*\blockheight +\blockheight -\blockheight/4 - \j*\blockheight/2 + \blockheight/2) {};
    }
    \foreach \c/\layers in {1/1, 2/2, 3/4, 4/8} {
        \def\x{\four+\c*\columnwidth - \columnwidth/2}
        \def\y{\blockheight-\b*\blockheight}
        \def\convradius{\blockheight/\layers}
        \foreach \n in {1, ..., \layers} {
            \ifthenelse{\c=4 \AND \n=3}{
                \def\yn{\y - \n*\convradius + 0.5*\convradius - \extrabufy}
            }{
                \ifthenelse{\c=4 \AND \n=5}{
                    \def\yn{\y - \n*\convradius + 0.5*\convradius - \extrabufy}
                }{
                    \ifthenelse{\c=4 \AND \n=7}{
                        \def\yn{\y - \n*\convradius + 0.5*\convradius - \extrabufy}
                    }{
                        \def\yn{\y - \n*\convradius + 0.5*\convradius}
                    }
                }
            }
            \def\xn{\x}
            \ifthenelse{\c=4}{
                \node[conv] (conv-\b-\c-\n) at (\xn, \yn) {};
            }{
                \node[noconv] (conv-\b-\c-\n) at (\xn, \yn) {};
            }
        }
    }

    \node (entry) at (\four+\columnwidth*\columns/2, 0.6) {};
    \node (below-entry) at (\four+\columnwidth*\columns/2, 0.1) {};

    \draw[noarrow,myround] (entry.south) to ($(entry.south) + (0, -0.45)$) to ($(entry.south) + (+\columnwidth*0.5, -0.45)$) to  (conv-\b-3-1);
    \draw[noarrow,myround] (entry.south) to ($(entry.south) + (0, -0.45)$) to ($(entry.south) + (-\columnwidth*1.5, -0.45)$) to  (conv-\b-1-1);
    \draw[noarrow,myround] (entry.south) to ($(entry.south) + (0, -0.45)$) to ($(entry.south) + (-\columnwidth*0.5, -0.45)$) to  (conv-\b-2-1);
    \draw[arrow,myround] (entry.south) to ($(entry.south) + (0, -0.45)$) to ($(entry.south) + (+\columnwidth*1.5, -0.45)$) to  (conv-\b-4-1);

    \draw[noarrow] (conv-\b-3-1) to ($(joiner-\b-3-1.north) - (\columnwidth/2, 0)$);
    \draw[noarrow,myround] (joiner-\b-3-1.south) to ($(joiner-\b-3-1.south) + (0, -\arrowbuf)$)  to ($(joiner-\b-3-1.south) + (-\columnwidth/2, -\arrowbuf)$)  to (conv-\b-3-2);
    \draw[noarrow] (conv-\b-3-2) to (joiner-\b-2-1);
    \draw[noarrow] (joiner-\b-2-1) to (conv-\b-3-3);
    \draw[noarrow] (conv-\b-3-3) to ($(joiner-\b-3-2.north) - (\columnwidth/2, 0)$);
    \draw[noarrow,myround] (joiner-\b-3-2.south) to ($(joiner-\b-3-2.south) + (0, -\arrowbuf)$) to ($(joiner-\b-3-2.south) + (-\columnwidth/2, -\arrowbuf)$) to (conv-\b-3-4);
    \draw[noarrow] (conv-\b-3-4) to ($(joiner-\b-1-1.north) + (\columnwidth*0.5, 0)$);

    \draw[noarrow] (conv-\b-2-1) to ($(joiner-\b-2-1.north) - (\columnwidth, 0)$);
    \draw[noarrow,myround] (joiner-\b-2-1.south) to ($(joiner-\b-2-1.south) + (0, -\arrowbuf)$) to ($(joiner-\b-2-1.south) + (-\columnwidth, -\arrowbuf)$) to (conv-\b-2-2);
    \draw[noarrow] (conv-\b-2-2) to ($(joiner-\b-1-1.north) + (-\columnwidth*0.5, 0)$);

    \draw[noarrow] (conv-\b-1-1) to ($(joiner-\b-1-1.north) - (\columnwidth*1.5, 0)$);

    \draw[arrow] (conv-\b-4-1) to (conv-\b-4-2);
    \draw[arrow] (conv-\b-4-2) to ($(joiner-\b-3-1.north) + (\columnwidth/2, 0)$);
    \draw[arrow,myround] (joiner-\b-3-1.south) to ($(joiner-\b-3-1.south) + (0, -\arrowbuf)$) to ($(joiner-\b-3-1.south) + (\columnwidth/2, -\arrowbuf)$) to (conv-\b-4-3);
    \draw[arrow] (conv-\b-4-3) to (conv-\b-4-4);
    \draw[arrow] (conv-\b-4-4) to ($(joiner-\b-2-1.north) + (\columnwidth, 0)$);
    \draw[arrow,myround] (joiner-\b-2-1.south) to ($(joiner-\b-2-1.south) + (0, -\arrowbuf)$) to ($(joiner-\b-2-1.south) + (\columnwidth, -\arrowbuf)$) to (conv-\b-4-5);
    \draw[arrow] (conv-\b-4-5) to (conv-\b-4-6);
    \draw[arrow] (conv-\b-4-6) to ($(joiner-\b-3-2.north) + (\columnwidth/2, 0)$);
    \draw[arrow,myround] (joiner-\b-3-2.south) to ($(joiner-\b-3-2.south) + (0, -\arrowbuf)$) to ($(joiner-\b-3-2.south) + (\columnwidth/2, -\arrowbuf)$) to (conv-\b-4-7);
    \draw[arrow] (conv-\b-4-7) to (conv-\b-4-8);
    \draw[arrow] (conv-\b-4-8) to ($(joiner-\b-1-1.north) + (\columnwidth*1.5, 0)$);

}

\node (exit) at (\four+\columnwidth*\columns/2, -\blocks*\blockheight-0.7) {\textbf{\textsf{\scriptsize{Iteration \#4}}}};
\node (exit-south) at (\four+\columnwidth*\columns/2, -\blocks*\blockheight-1.0) {\textbf{\textsf{\tiny{(Global)}}}};
\draw[arrow] (joiner-1-1-1) to  (exit);

\begin{pgfonlayer}{background}
\filldraw [line width=4mm,join=round,black!10]
  (below-entry.south -| conv-1-4-1.east) rectangle (joiner-1-1-1.south -| conv-1-1-1.west);
\end{pgfonlayer}

\end{tikzpicture}

%% file: include/sec4-results.tex
\begin{figure}
   \input{include/table_all}
   \captionof{table}{
      \textbf{CIFAR-100/CIFAR-10/SVHN.}
      We compare test error (\%) with other leading methods, trained with
      either no data augmentation, translation/mirroring (+), or more
      substantial augmentation (++).  Our main point of comparison is
      {\resnet}.  We closely match its benchmark results using data
      augmentation, and outperform it by large margins without data
      augmentation.  Training with {\droppath}, we can extract from
      {\fracnet} single-column (plain) networks that are highly competitive.
   }
\label{tab:all}
\end{figure}

\addtocounter{footnote}{-1}
\footnotetext{%
   Densely connected networks (DenseNets) are concurrent work, appearing
   subsequent to our original arXiv paper on FractalNet.  A variant of residual
   networks, they swap addition for concatenation in the residual functional
   form.  We report performance of their $250$-layer DenseNet-BC network with
   growth rate $k=24$.
}

\stepcounter{footnote}
\footnotetext{%
   This deeper (4 column) {\fracnet} has fewer parameters.  We vary column
   width: $(128, 64, 32, 16)$ channels across columns initially, doubling each
   block except the last.  A linear projection temporarily widens thinner
   columns before joins.  As in~\cite{SqueezeNet}, we switch to a mix of
   $1 \times 1$ and $3 \times 3$ convolutional filters.
}

The CIFAR, SVHN, and ImageNet datasets serve as testbeds for comparison to
prior work and analysis of {\fracnet}'s internal behavior.  We evaluate
performance on the standard classification task associated with each
dataset.  For CIFAR and SVHN, which consist of $32 \times 32$ images, we set
our fractal network to have $5$ blocks ($B=5$) with $2 \times 2$
non-overlapping max-pooling and subsampling applied after each.  This reduces
the input $32 \times 32$ spatial resolution to $1 \times 1$ over the course of
the entire network.  A softmax prediction layer attaches at the end of the
network.  Unless otherwise noted, we set the number of filter channels within
blocks $1$ through $5$ as $(64, 128, 256, 512, 512)$, mostly matching the
convention of doubling the number of channels after halving spatial resolution.

For ImageNet, we choose a fractal architecture to facilitate direct comparison
with the $34$-layer {\resnet} of~\cite{he2015deep}.  We use the same first and
last layer as {\resnet}-34, but change the middle of the network to consist of
$4$ blocks ($B=4$), each of $8$ layers ($C=4$ columns).  We use a filter
channel progression of $(128, 256, 512, 1024)$ in blocks $1$ through $4$.

\subsection{Training}
\label{sec:training}

For experiments using {\dropout}, we fix drop rate per block at
$(0\%, 10\%, 20\%, 30\%, 40\%)$, similar to~\cite{elu}.  Local {\droppath}
uses $15\%$ drop rate across the entire network.

We run for $400$ epochs on CIFAR, $20$ epochs on SVHN, and $70$ epochs on
ImageNet.  Our learning rate starts at $0.02$ (for ImageNet, $0.001$) and we
train using stochastic gradient descent with batch size $100$ (for ImageNet,
$32$) and momentum $0.9$.  For CIFAR/SVHN, we drop the learning rate by a
factor of $10$ whenever the number of remaining epochs halves.  For ImageNet,
we drop by a factor of $10$ at epochs $50$ and $65$.  We use Xavier
initialization~\citep{glorot2010understanding}.

A widely employed~\citep{nin,elu,srivastava2015highway,he2015deep,
he2016identity,huang2016stochasticdepth,rir} scheme for data augmentation on
CIFAR consists of only horizontal mirroring and translation (uniform offsets
in $[-4,4]$), with images zero-padded where needed after mean subtraction.
We denote results achieved using no more than this degree of augmentation by
appending a ``+'' to the dataset name (\eg~CIFAR-100+).  A ``++'' marks
results reliant on more data augmentation; here exact schemes may vary.  Our
entry in this category is modest and simply changes the zero-padding to
reflect-padding.

\begin{figure}
   \begin{minipage}[b]{0.45\linewidth}
      \vspace{0pt}
      \begin{center}
      \input{include/table_imagenet}
      \end{center}
      \vspace{-0.04\linewidth}
      \captionof{table}{
         \textbf{ImageNet}~(validation set, 10-crop).
      }
      \label{tab:imagenet}
      \vspace{0.03\linewidth}
      \begin{center}
      \input{include/table_columns}
      \end{center}
      \vspace{-0.04\linewidth}
      \captionof{table}{
         \textbf{Ultra-deep fractal networks} (CIFAR-100++).
         Increasing depth greatly improves accuracy until eventual diminishing
         returns.  Contrast with plain networks, which are not trainable if
         made too deep (Table~\ref{tab:individual}).
      }
      \label{tab:columns}
   \end{minipage}
   \hfill
   \begin{minipage}[b]{0.51\linewidth}
      \vspace{0pt}
      \begin{center}
      \input{include/table_individual}
      \end{center}
      \vspace{-0.02\linewidth}
      \captionof{table}{
         \textbf{Fractal structure as a training apparatus} (CIFAR-100++).
         Plain networks perform well if moderately deep, but exhibit worse
         convergence during training if instantiated with great depth.
         However, as a column trained within, and then extracted from, a
         fractal network with mixed {\droppath}, we recover a plain network
         that overcomes such depth limitation (possibly due to a
         student-teacher effect).
      }
      \label{tab:individual}
   \end{minipage}
\end{figure}

\subsection{Results}
\label{sec:evaluation}

Table~\ref{tab:all} compares performance of {\fracnet} on CIFAR and SVHN with
competing methods.  {\Fracnet} (depth $20$) outperforms the original {\resnet}
across the board.  With data augmentation, our CIFAR-100 accuracy is close to
that of the best {\resnet} variants.  With neither augmentation nor
regularization, {\fracnet}'s performance on CIFAR is superior to both {\resnet}
and {\resnet} with stochastic depth, suggesting that {\fracnet} may be less
prone to overfitting.  Most methods perform similarly on SVHN.  Increasing
depth to $40$, while borrowing some parameter reduction tricks~\citep{
SqueezeNet}, reveals {\fracnet}'s performance to be consistent across a range
of configuration choices.

Experiments without data augmentation highlight the power of {\droppath}
regularization.  On CIFAR-100, {\droppath} reduces {\fracnet}'s error rate
from $35.34\%$ to $28.20\%$.  Unregularized {\resnet} is far behind ($44.76\%$)
and {\resnet} with stochastic depth ($37.80\%$) does not catch up to our
unregularized starting point of $35.34\%$.  CIFAR-10 mirrors this story.  With
data augmentation, {\droppath} provides a boost (CIFAR-10), or does not
significantly influence {\fracnet}'s performance (CIFAR-100).

Note that the performance of the deepest column of the fractal network is
close to that of the full network (statistically equivalent on CIFAR-10).
This suggests that the fractal structure may be more important as a
learning framework than as a final model architecture.

Table~\ref{tab:imagenet} shows that {\fracnet} scales to ImageNet,
matching {\resnet}~\citep{he2015deep} at equal depth.  Note that, concurrent
with our work, refinements to the residual network paradigm further improve
the state-of-the-art on ImageNet.  Wide residual networks~\citep{wideresnet}
of $34$-layers reduce single-crop Top-1 and Top-5 validation error by
approximately $2\%$ and $1\%$, respectively, over ResNet-$34$ by doubling
feature channels in each layer.  DenseNets~\citep{densenet} substantially
improve performance by building residual blocks that concatenate rather than
add feature channels.

Table~\ref{tab:columns} demonstrates that {\fracnet} resists performance
degradation as we increase $C$ to obtain extremely deep networks ($160$ layers
for $C=6$).  Scores in this table are not comparable to those in
Table~\ref{tab:all}.  For time and memory efficiency, we reduced block-wise
feature channels to $(16, 32, 64, 128, 128)$ and the batch size to $50$ for the
supporting experiments in Tables~\ref{tab:columns} and~\ref{tab:individual}.

Table~\ref{tab:individual} provides a baseline showing that training of plain
deep networks begins to degrade by the time their depth reaches $40$ layers.
In our experience, a plain $160$-layer completely fails to converge. This
table also highlights the ability to use {\fracnet} and {\droppath} as an
engine for extracting trained networks (columns) with the same topology as
plain networks, but much higher test performance.

\begin{figure}
   \begin{center}
      \includegraphics[width=1.0\linewidth]{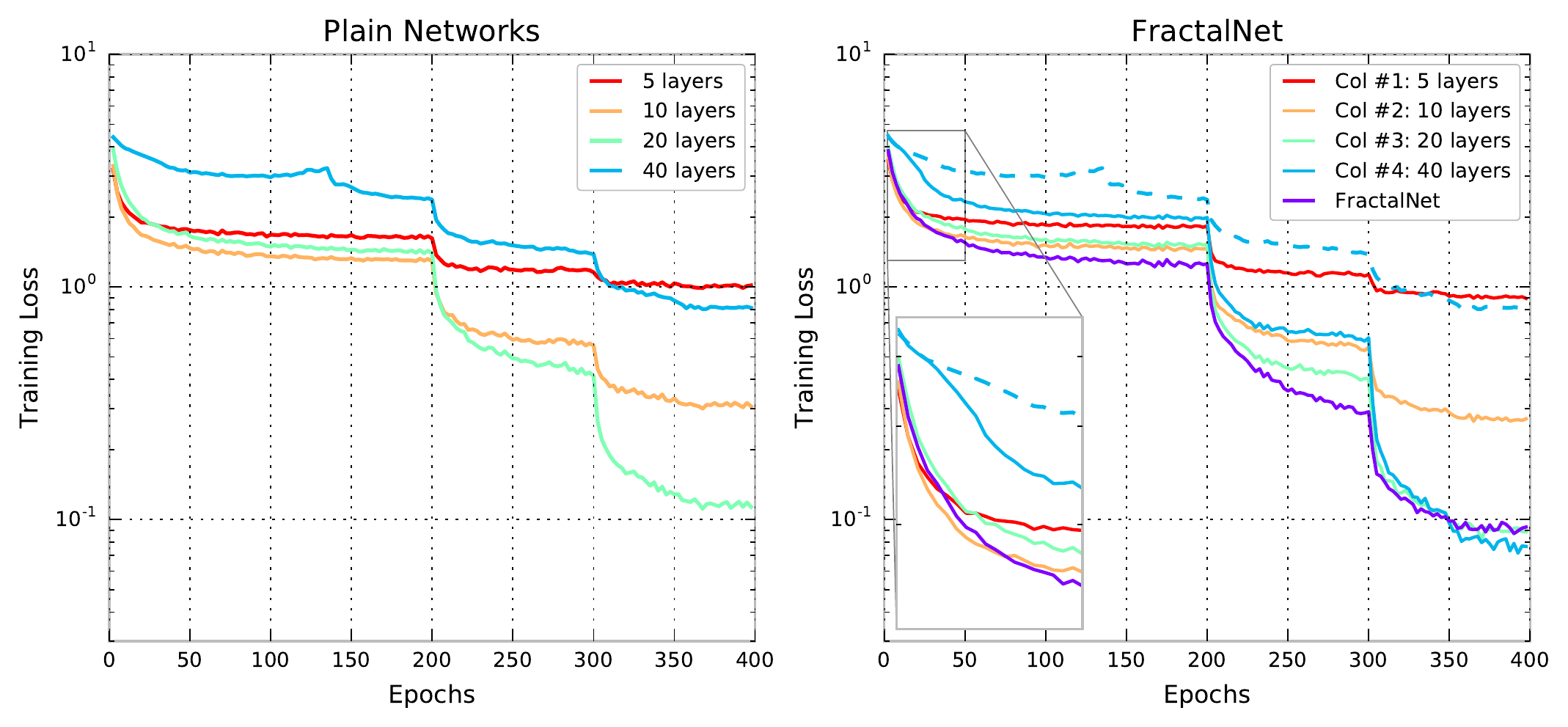}
   \end{center}
   \vspace{-0.4cm}
   \caption{
      \textbf{Implicit deep supervision.}
      \emph{Left:}
         Evolution of loss for plain networks of depth $5$, $10$, $20$ and
         $40$ trained on CIFAR-100.  Training becomes increasingly difficult
         for deeper networks.  At $40$ layers, we are unable to train the
         network satisfactorily.
      \emph{Right:}
         We train a $4$ column fractal network with mixed \droppath{},
         monitoring its loss as well as the losses of its four subnetworks
         corresponding to individual columns of the same depth as the plain
         networks.  As the $20$-layer subnetwork starts to stabilize,
         \droppath{} puts pressure on the $40$-layer column to adapt, with the
         rest of the network as its teacher.  This explains the elbow-shaped
         learning curve for Col \#4 that occurs around $25$ epochs.
   }
\label{fig:fractalnet-loss}
\end{figure}

\subsection{Introspection}
\label{sec:introspection}

With Figure~\ref{fig:fractalnet-loss}, we examine the evolution of a $40$-layer
{\fracnet} during training.  Tracking columns individually (recording their
losses when run as stand-alone networks), we observe that the $40$-layer column
initially improves slowly, but picks up once the loss of the rest of the
network begins to stabilize.  Contrast with a plain $40$-layer network trained
alone (dashed blue line), which never makes fast progress.  The column has the
same initial plateau, but subsequently improves after $25$ epochs, producing a
loss curve uncharacteristic of plain networks.

We hypothesize that the fractal structure triggers effects akin to deep
supervision and lateral student-teacher information flow.  Column \#4 joins
with column \#3 every other layer, and in every fourth layer this join
involves no other columns.  Once the fractal network partially relies on the
signal going through column \#3, {\droppath} puts pressure on column \#4 to
produce a replacement signal when column \#3 is dropped.  This task has
constrained scope.  A particular drop only requires two consecutive layers in
column \#4 to substitute for one in column \#3 (a mini student-teacher
problem).

This explanation of FractalNet dynamics parallels what, in concurrent
work, \cite{unrolled} claim for ResNet.  Specifically, \cite{unrolled} suggest
residual networks learn unrolled iterative estimation, with each layer
performing a gradual refinement on its input representation.  The deepest
FractalNet column could behave in the same manner, with the remainder of the
network acting as a scaffold for building smaller refinement steps by doubling
layers from one column to the next.

These interpretations appear not to mesh with the conclusions of~\cite{
veit16residual}, who claim that ensemble-like behavior underlies the success of
ResNet.  This is certainly untrue of some very deep networks, as FractalNet
provides a counterexample: we can extract a single column (plain network
topology) and it alone (no ensembling) performs nearly as well as the entire
network.  Moreover, the gradual refinement view may offer an alternative
explanation for the experiments of~\cite{veit16residual}.  If each layer makes
only a small modification, removing one may look, to the subsequent portion of
the network, like injecting a small amount of input noise.  Perhaps noise
tolerance explains the gradual performance degradation that \cite{
veit16residual} observe when removing ResNet layers.

%% file: include/table_all.tex
\def\dash{-\phantom{00}}
\begin{small}
\begin{tabular}{@{}l|r:rr|r:rr|r}
Method                                                   & C100  & C100+ & C100++ & C10   & C10+  & C10++ & SVHN \\
\toprule
Network in Network~\citep{nin}                           & 35.68 & \dash & \dash  & 10.41 & 8.81  & \dash & 2.35 \\
Generalized Pooling~\citep{lee2016generalizing}          & 32.37 & \dash & \dash  & 7.62  & 6.05  & \dash & 1.69 \\ 
Recurrent CNN~\citep{liang2015recurrent}                 & 31.75 & \dash & \dash  & 8.69  & 7.09  & \dash & 1.77 \\ 
Multi-scale~\citep{liao2015competitive}                  & 27.56 & \dash & \dash  & 6.87  & \dash & \dash & 1.76 \\
FitNet~\cite{romero2014fitnets}                          & \dash & 35.04 & \dash  & \dash & 8.39  & \dash & 2.42 \\ 
Deeply Supervised~\citep{lee2014deeply}                  & \dash & 34.57 & \dash  & 9.69  & 7.97  & \dash & 1.92 \\
All-CNN~\citep{springenberg2014striving}                 & \dash & 33.71 & \dash  & 9.08  & 7.25  & 4.41  & \dash\\ 
Highway Net~\citep{srivastava2015highway}                & \dash & 32.39 & \dash  & \dash & 7.72  & \dash & \dash\\ 
ELU~\citep{elu}                                          & \dash & 24.28 & \dash  & \dash & 6.55  & \dash & \dash\\ 
Scalable BO~\citep{snoek2015scalable}                    & \dash & \dash & 27.04  & \dash & \dash & 6.37  & 1.77 \\ 
Fractional Max-Pool~\citep{graham2014fractional}         & \dash & \dash & 26.32  & \dash & \dash & 3.47  & \dash\\ 
\midrule
FitResNet~\citep{mishkin2015all}                         & \dash & 27.66 & \dash  & \dash & 5.84  & \dash & \dash\\ 
{\Resnet}~\citep{he2015deep}                             & \dash & \dash & \dash  & \dash & 6.61  & \dash & \dash\\ 
{\Resnet} by~\citep{huang2016stochasticdepth}            & 44.76 & 27.22 & \dash  & 13.63 & 6.41  & \dash & 2.01 \\ 
Stochastic Depth~\citep{huang2016stochasticdepth}        & 37.80 & 24.58 & \dash  & 11.66 & 5.23  & \dash & 1.75 \\
Identity Mapping~\citep{he2016identity}                  & \dash & 22.68 & \dash  & \dash & 4.69  & \dash & \dash\\
{\Resnet} in {\Resnet}~\citep{rir}                       & \dash & 22.90 & \dash  & \dash & 5.01  & \dash & \dash\\    
Wide~\citep{wideresnet}                                  & \dash & 20.50 & \dash  & \dash & 4.17  & \dash & \dash\\ 
DenseNet-BC~\citep{densenet}\footnotemark                & 19.64 & 17.60 & \dash  &  5.19 & 3.62  & \dash & 1.74 \\
\midrule
{\Fracnet} (20 layers, 38.6M params)                     & 35.34 & 23.30 & 22.85  & 10.18 & 5.22  & 5.11  & 2.01 \\ 
~+~\droppath~+~\dropout                                  & 28.20 & 23.73 & 23.36  & 7.33  & 4.60  & 4.59  & 1.87 \\ 
\quad\quad%
\raisebox{0.06cm}{$\drsh$} deepest column alone          & 29.05 & 24.32 & 23.60  & 7.27  & 4.68  & 4.63  & 1.89 \\
{\Fracnet} (40 layers, 22.9M params)\footnotemark        & \dash & 22.49 & 21.49  & \dash & 5.24  & 5.21  & \dash\\ 
\bottomrule
\end{tabular}
\end{small}
%

%% file: include/table_imagenet.tex
\begin{tabular}{@{}lrr@{}}
Method            & Top-1 (\%) & Top-5  (\%) \\
\toprule
VGG-16            & 28.07      & 9.33 \\
{\Resnet}-34 C    & 24.19      & 7.40 \\
{\Fracnet}-34     & 24.12      & 7.39 \\
\bottomrule
\end{tabular}

%% file: include/table_columns.tex
\begin{tabular}{@{}lrrr@{}}
  Cols. & Depth  & Params.    &  Error (\%)\\
\toprule
     ~1 &      5 &       0.3M &      37.32 \\   
     ~2 &     10 &       0.8M &      30.71 \\   
     ~3 &     20 &       2.1M &      27.69 \\   
     ~4 &     40 &       4.8M &      27.38 \\   
     ~5 &     80 &      10.2M &      26.46 \\   
     ~6 &    160 &      21.1M &      27.38 \\   
\bottomrule
\end{tabular}

%% file: include/table_individual.tex
\begin{tabular}{@{}lrrr@{}}
Model                    & Depth          & Train Loss     & Error (\%)     \\
\toprule
Plain                    & 5              & 0.786          & 36.62          \\
Plain                    & 10             & 0.159          & 32.47          \\
Plain                    & 20             & 0.037          & 31.31          \\
\textbf{Plain}           & \textbf{40}    & \textbf{0.580} & \textbf{38.84} \\
\midrule
Fractal Col \#1          & 5              & 0.677          & 37.23          \\
Fractal Col \#2          & 10             & 0.141          & 32.85          \\
Fractal Col \#3          & 20             & 0.029          & 31.31          \\
\textbf{Fractal Col \#4} & \textbf{40}    & \textbf{0.016} & \textbf{31.75} \\
\midrule
\textbf{Fractal Full}    & \textbf{40}    & \textbf{0.015} & \textbf{27.40} \\
\bottomrule
\end{tabular}
%
%
%

%% file: include/sec5-final.tex
Our experiments with fractal networks provide strong evidence that path length
is fundamental for training ultra-deep neural networks; residuals are
incidental.  Key is the shared characteristic of {\fracnet} and {\resnet}:
large nominal network depth, but effectively shorter paths for gradient
propagation during training.  Fractal architectures are arguably the simplest
means of satisfying this requirement, and match residual networks
in experimental performance.  Fractal networks are resistant to being too deep;
extra depth may slow training, but does not impair accuracy.

With {\droppath}, regularization of extremely deep fractal networks is
intuitive and effective.  {\Droppath} doubles as a method of enforcing speed
(latency) vs.~accuracy tradeoffs.  For applications where fast responses have
utility, we can obtain fractal networks whose partial evaluation yields good
answers.

Our analysis connects the internal behavior of fractal networks with phenomena
engineered into other networks.  Their substructure resembles hand-crafted
modules used as components in prior work.  Their training evolution may emulate
deep supervision and student-teacher learning.